%% file: conference_101719.tex
\documentclass[letterpaper, 10pt, conference]{ieeeconf}
\IEEEoverridecommandlockouts
\pdfoutput=1

\usepackage{balance}
\usepackage{url}
\usepackage{epsfig}
\usepackage{multirow}
\usepackage{graphicx}
\usepackage{graphics}
\usepackage{xcolor}

\title{\LARGE \bf
A ROS Architecture for Personalised HRI\\ with a Bartender Social Robot*}

\author{Alessandra Rossi$^{1}$, Maria Di Maro$^{2}$, Antonio Origlia$^{1}$, Agostino Palmiero$^{3}$ and Silvia Rossi$^{1}$% <-this % stops a space
\thanks{*This work has been supported by Italian PON I\&C 2014-2020 within the BRILLO research project ``Bartending Robot for Interactive Long-Lasting Operations'', no. F/190066/01-02/X44.}% <-this % stops a space
\thanks{$^{1}$Alessandra Rossi, Antonio Origlia and Silvia Rossi are with the Department of Electrical Engineering and Information Technologies, University of Naples Federico II, Napoli, Italy {\tt\small \{alessandra.rossi, antonio.origlia, silvia.rossi\}@unina.it}}%
\thanks{$^{2}$Maria Di Maro is with the Interdepartmental Center for Advances in Robotic Surgery, University of Naples Federico II, Naples, Italy {\tt\small maria.dimaro2@unina.it}}%
\thanks{$^{3}$Agostino Palmiero is with Totaro Automazioni s.r.l., Marcianise, Italy {\tt\small a.palmiero@totaroautomazioni.it}}%
}

\begin{document}

\maketitle
\thispagestyle{empty}
\pagestyle{empty}

\begin{abstract}
BRILLO (Bartending Robot for Interactive Long-Lasting Operations) project has the overall goal of creating an autonomous robotic bartender that can interact with customers while accomplishing its bartending tasks. In such a scenario, people's novelty effect connected to the use of an attractive technology is destined to wear off and, consequently, it negatively affects the success of the service robotics application. For this reason, providing personalised natural interaction while accessing its services is of paramount importance for increasing users' engagement and, consequently, their loyalty.
In this paper, we present the developed three-layers ROS architecture integrating a perception layer managing the processing of different social signals, a decision-making layer for handling multi-party interactions, and an execution layer controlling the behaviour of a complex robot composed of arms and a face. Finally, user modelling through a beliefs layer allows for personalized interaction. 
%The development of such complex HRI s a challenge faced by many in the Human-Robot Interaction (HRI) community.
%into a real-world multi-party interactive scenario. % is discussed along with its personalization of the interaction. 
%Such a robot will be able to adapt its behaviours to customers' requests, preferences, and needs with the aim of increasing customers' engagement through natural dialogues and situational awareness.
\end{abstract}

%\begin{IEEEkeywords}
%Personalised HRI, service robotics, multi-modal interaction, long-lasting collaborations, robot bartender
%\end{IEEEkeywords}

\section{INTRODUCTION}
\label{sec:intro}
\input{sections/intro.tex}

%\section{Related Works}
%\label{sec:intro}
%\input{sections/related.tex}

\section{MULTI-MODAL AND MULTI-USERS SCENARIOS}
\label{sec:scenario}
\input{sections/scenario.tex}

\subsection{BRILLO Bartender Robot}
\label{sec:hard}
\input{sections/hardware.tex}

\section{BRILLO ROS ARCHITECTURE}
\label{sec:sys}
\input{sections/brillo.tex}

\section{FUTURE WORKS / CONCLUSIONS}
\label{sec:conc}
\input{sections/conc.tex}
\bibliographystyle{IEEEtran}
\balance 
\bibliography{biblio}
\end{document}

%% file: sections/intro.tex
%Composing HARMONI: An Open-source Tool for Human and Robot Modular OpeN Interaction
%ROS for Human-Robot Interaction

In recent years, service robots have been employed in a variety of contexts that involve direct interactions with multi-users in public environments. A particularly challenging one is the bartending domain that combines the complexity of efficiently manipulating objects and the need of keeping users engaged for a long-lasting interaction. This, particularly, reflects the approaches of modern businesses that aim to achieve customers' satisfaction and retail by presenting an equally high quality product and service \cite{10.1145/3450614.3463423}.
Several projects \cite{1626700,8595543} explored the robotic bartending domain by developing automatic serving robots that are able to serve multiple users. 
Long-lasting interactions, however, can be established when the service robot is capable of showing social intelligence through personalised interactions \cite{Rossi2020}. %Conversely, non-socially credible robots are more likely to be perceived negatively during human-robot interaction (HRI). %\cite{9223471}. 

Current literature has identified several aspects that affect people's perception of social intelligence in robots. For example, a robot with facial features can be perceived as more intelligent than one without any \cite{Walters2009}, a robot that is able to move naturally can enhance people's acceptance of the robot and convey a sense of security \cite{Lichtenthaler2012}, and a robot that is able to model human behaviours and express appropriate emotions can positively affect the interactions \cite{Wykowska2014}. Among those, the possibility of providing personalised services can increase users' interaction on a long-timescale and their involvement with the robot \cite{10.1108/IJCHM-09-2016-0520}. Socially intelligent bartender robots that are able to mix task execution, dialogue, and social interaction in response to customers' states and intentions were more efficient than non-social ones \cite{ 10.1145/2522848.2522879}. %The integration of a robotic system that can combine a ``system of record'' (e.g., analysing and storing past interactions, preferences to optimise sales with respect to the specific users) and a ``systems of engagement'' (e.g., aiming at facilitating and enhancing the experience via a personalised and natural interaction) plays a central role in customer relationship management. 
A robotic system that integrates a ``system of record'' (e.g., analysing and storing past interactions, preferences to optimise sales with respect to the specific users) and a ``system of engagement'' (e.g., aiming at facilitating and enhancing the experience via a personalised and natural interaction) can play a crucial role in customers relationship management.

To achieve this goal, 
%Typical approaches for the development of robots are represented by cognitive architectures \cite{LIETO20181}. 
a complex human-robot interaction (HRI) and control architecture have to be designed that comprehend different software components allowing for efficient and simultaneous execution of multiple tasks and for providing essential capabilities, such as storing past events \cite{prescott2019memory}, constructing models of others' actions, beliefs, desires, and intentions \cite{LEMAIGNAN201745}, modelling the domain knowledge, selecting actions and behaviours, and planning \cite{moulin2017dac}. For example, in the JAMES project \cite{10.1145/2522848.2522879}, a bartender robot was able to engage participants in conversation producing facial expressions and lip-synchronising speech. 
The iCub robot in Tanevska et al.'s study \cite{8850688} adapted its gaze and body to convey comfort and discomfort according to the engagement of the participants. The CORTEX cognitive architecture \cite{7101621} was used to allow a salesman robot to convince potential customers to follow the robot towards a selling boot. The robot was able to identify the customers, understand people's willingness of following it, and answer some specific questions. 

While cognitive architectures have been investigated for a long time, real social and service robot implementations in complex scenarios, such as the bartending service, have only recently been developed \cite{Vicente2015}. Moreover, there is an inconsistency between current robots' ability to generate verbal and non-verbal expressive behaviours (such as spoken language, gestures, emotions), and their capability of understanding the situational context and engaging the users in natural dialogues according to the users' intentions and desires \cite{PRESCOTT2021101993}. The BRILLO project aims to create a robotic platform that is able to accomplish both the expected management of a bar counter, such as drinks manipulation, and the socially intelligent interaction, context-awareness, and personalisation. The development of such interaction and personalization capabilities in real-world settings is still an open challenge for the HRI community due to the complexity of interaction pipelines and the lack of maturity of some of the underlying detection and processing algorithms \cite{mohamed2021ros4hri}.   
%In this work, we present a description of architecture developed for the BRILLO project.%, its framework, and the applicative scenarios considering multiple-users and their preferences, moods, and differences.
%In this paper, we present a description of the BRILLO project and its framework, and the applicative user scenarios considering the needs of the multiple agents involved in the interaction, in terms of preferences, moods, and differences.

%% file: sections/scenario.tex
A typical use case scenario of a BRILLO service point includes a human user, a bartender robot for preparing and serving drinks, and a totem kiosk to register, recognise the users and manage orders.
%\subsubsection{Scenarios}
Users' first interactions are with the totem kiosk %(see Figure \ref{fig:use_case_1})
that welcomes them and allows them to register in case it is their first visit. If the users are returning customers, they are recognised via bio-metric face recognition. Then, the customers are able to order a drink from the menu displayed on the kiosk or they can decide to make their order sitting at the counter by directly interacting with the bartender robot. 
At the bar station, %(see Figure \ref{fig:use_case_2})
the bartender robot recognises the registered users, takes orders (in case they are not placed at the kiosk), and serves them. The bartender robot interacts with the users according to their profile which is built upon typical customers' personas (such as workers on a lunch break, groups of friends or family members, and regular users) and needs, such as previous orders, knowledge about the user's general interests, observing their engagement levels, processing their moods based on the sentiment analysis of their dialogues and facial expressions. The bartender robot is also able to manage multiple orders and users, by opportunely scheduling and adapting its behaviours.

The richness of the one-to-one and one-to-many human-robot interactions in the above-mentioned scenarios requires the development of a complex and sophisticated HRI architecture that allows the robot to show a social comprehension of the context and other agents involved in the interaction, and, at the same time, generates matching verbal and non-verbal socially acceptable behaviours. The bartender robot, for example, needs to intelligently adapt its dialogues, pose and gestures, according to the user's needs, in terms of situational context (drink orders, group dynamics, etc.).

%% file: sections/hardware.tex
We adopted a minimalist anthropomorphic structure for the bartender robot (see Figure \ref{fig:syslayers}). The robot has two Kuka\footnote{Kuka Robotics \url{https://www.kuka.com}} LBR iiwa 14 R820 robotic arms (each of 7 DoF and gripper), attached to a fixed-torso (only one arm has been integrated into the current version of the robot), and a Furhat Robotics head\footnote{Furhat Robotics \url{https://furhatrobotics.com}} (called Furhat, 3 DoF). %It is possible to change the Furhat robot's face features, including gender, age and ethnicity, and the voice and language to match the persona created. 
The bartender robot is equipped with a variety of external sensors to improve and support its capability to perceive and assess the environment, the users, and the activities of the other agents involved in the interaction. In particular, a 4x2MP IR 180$^{\circ}$ Multi-sensor Panoramic Network Bullet camera\footnote{https://us.dahuasecurity.com/?product=4x2mp-ir-180-multi-sensor-panoramic-network-bullet} is mounted under the furhat head and two microphone arrays\footnote{PureAudio USB Array Microphone} are placed on the right and on the left of the robot to perform source separation and noise reduction to isolate the customers’ voices.

Currently, this robot is able to prepare smoothies and cocktails.
The choice to adopt an anthropomorphic structure for the robot comes from the evidence that people are social entities, and they are more comfortable while interacting with agents that can show social behaviours \cite{978-3-030-23807-0}. %More specifically, people's perception of a robot's social intelligence can be affected by the robot's anthropomorphic appearance, its level of autonomy, and its functionalities \cite{Staffa2016200}. 

\subsection{Interaction Patterns and Dialogue}
The foreseen interaction pattern with the bartender robot is composed of the following phases: 1) greetings or greetings and wait; 2) recommendation; 3) orders and changes request; 4) order confirmation; 5) personalised casual interaction; 6) complimentary close. In the presence of a single user, the robot welcomes them (``greetings") and, then, recommends drinks, while in the presence of multiple users, the robot welcomes them, and invites them to wait their turn to be served (``greetings and wait"). After the recommendation of the drinks phase, the customer can place their order and/or ask for order customisation (``orders and changes request"). If no other change is needed, the order can be confirmed to the robot (``order confirmation"). 
During the drink preparation phase, the robot can converse with the user to entertain them. The robot engages the customer in casual dialogues based on data collected in previous interactions. According to the social feedback received from past interactions or previous turns, appropriate topics of conversation are selected. 
%During the interaction, the input is provided by the \textit{Context-awareness and Decision-making} modules.% by analysing the customer's focus of attention, mood, and sentiment analysis (``personalised casual interaction"). This capability also functions 
%as preparation time filler and user engagement support. 
The robot closes the interaction after it prepared the drink (``complimentary close").

%\subsubsection{The waiter robot}

%While the bartender robot needs to engage the users in conversations, the main purpose of the waiter robot is to serve the drinks at the table. In such scenario, the robot needs to be able to adopt socially acceptable navigational and approaching skills, such as regulating its velocity, navigating in cluttered environments without disturbing or colliding with people and objects, and minding people's personal spaces \cite{978-3-030-23807-0,pacchierotti2006,Koay2007,Walters2011}. The waiter robot is equipped with a serving tray, multi-directional wheels and a several sensors to autonomously and safely navigate in the environment, and serve the drinks to the users.

%\subsubsection{Totem}
%The totem used to welcome users upon their arrival is equipped with a camera to detect and recognise new or returning customers. The totem allows new users to register at their first use, provide preferences and personal information to be used to personalise their experience, and making and managing their drink orders. Each table of the BRILLO service establishment has a common tablet allowing customers to make and manage orders that are prepared by the bartender robot and brought by the waiter robot. 

%% file: sections/brillo.tex
\begin{figure}[!t]
  \centering
  \includegraphics[width=\linewidth]{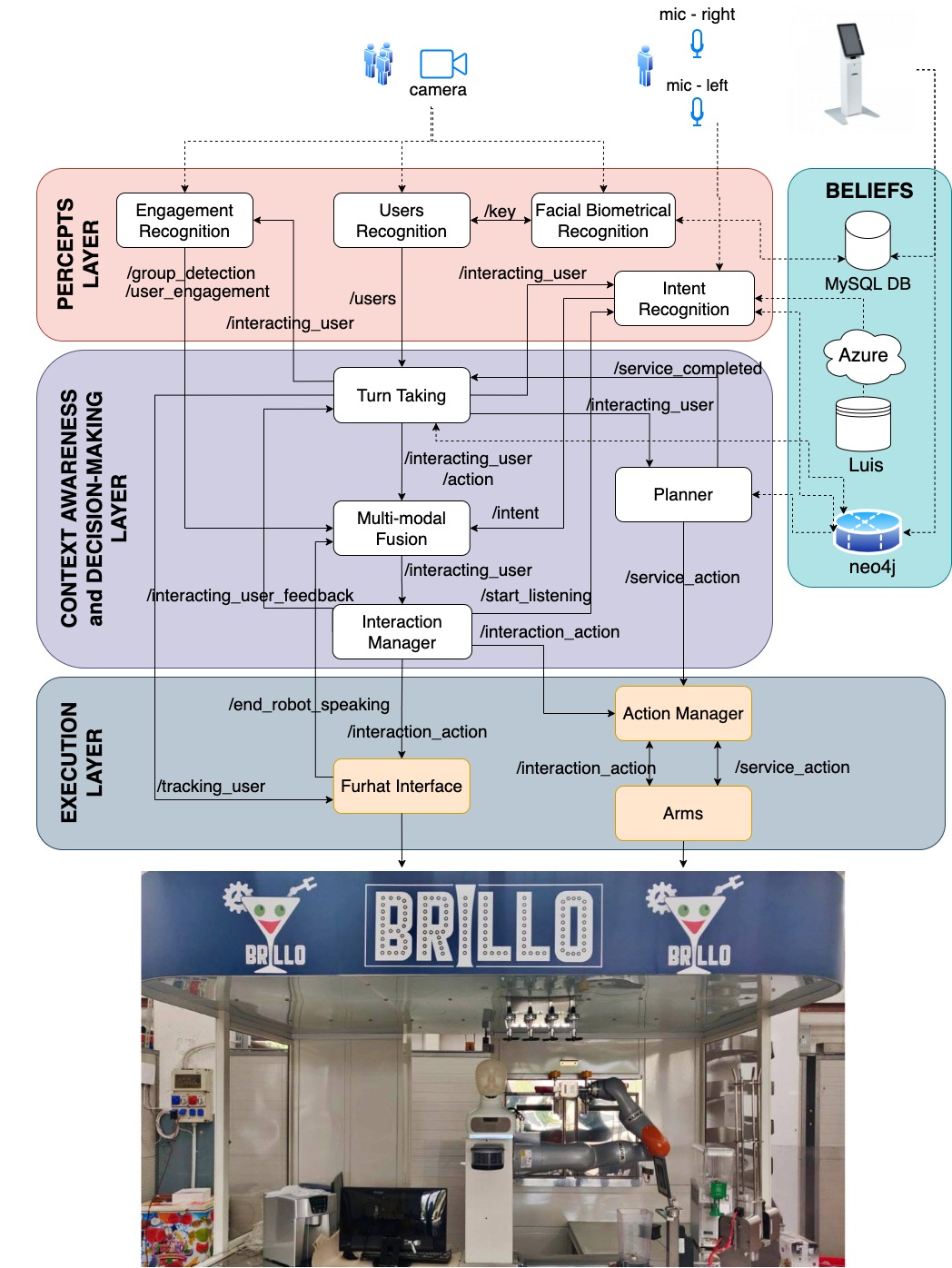}
  \caption{Overview of the BRILLO project's ROS architecture and topics. It is composed of a component to build the Beliefs and three principal layers: 1) Percepts, 2) Context-awareness and Decision Making, and 3) Execution. %Each layer has a collection of different modules.
  }
  \label{fig:syslayers}
\end{figure}

In a recent survey on cognitive architectures, Kotseruba and Tsotsos \cite{Tsotsos} identified seven core cognitive abilities - perception, attention mechanisms (used in multi-user interaction), action selection, memory, preference learning, reasoning, and metareasoning. 
We started from such core cognitive abilities (excluding metareasoning) to develop the functionalities of the BRILLO system to enable the robot 
%involved in the interaction 
to perceive the users, build dynamic and efficient interactions while adapting the robots' behaviours to the users' needs and preferences. 

%The next sections describe the cognitive architecture adopted for the BRILLO project.
%The high-level design of the architecture is inspired by the BDI (Belief-Desire-Intention) model \cite{10.1007/BFb0026757}. This type of design allows for modelling both reactivity and pro-activity in the robot behaviours, interaction with multi-users by means of multiple instantiated intentions, while modelling long-term knowledge as well as situational assessment through belief update mechanisms. 

The global architecture adopted for the BRILLO project is shown in Figure \ref{fig:syslayers}.
%
%\subsection{Architecture}
From a software engineering perspective, the developed modules operate asynchronously by means of ROS nodes and communication via topic subscriptions.

%Here discuss the BDI structure of the modules.

%Figure \ref{fig:syslayers} presents an overview of the architecture and relevant implemented modules. This architecture is based on the same principles of common layered architectures (such as \cite{ENRICHME2020,LEMAIGNAN201745,Burattini2012129}). This project aims to develop a robotic system that is able to work as bartender, completing multiple and in parallel tasks while adapting its behaviours to its users' needs and personal differences, in terms of preferences, moods and traits.

%According to the current literature that confirms the importance of verbal and non-verbal communication following a multimodal approach and supplemented by social skills for the improvement of service quality and loyalty, BRILLO's architecture will follow a user-centered approach. In particular, the design of the interfaces for accessing the functions of the bartending robot will be divided into a functional part, dedicated to identifying the services on which to evaluate performance, and a social part, dedicated to specifying the behaviours to be implemented in relation to the different user profiles.

%Our system is a four-layers control architecture where each component is linked to the Beliefs, Desires, Intentions (BDI) model architectures \cite{LEMAIGNAN201745,ENRICHME2020}. 

\subsection{Percepts Layer}
\label{subsec:perc}
\input{sections/percept}

\subsection{Users' Beliefs}
\label{subsec:beliefs}
\input{sections/beliefs}

\subsection{Context Awareness and Decision Making Layer}
\label{subsec:awareness} 
\input{sections/awareness}

\subsection{Execution Layer}
\label{subsec:exec}
\input{sections/exec}

%% file: sections/percept.tex
The \textit{Percepts layer} manages the robots' multi-modal perception capabilities which consist of the information obtained by the modules for processing the visual and speech inputs. %Other perceptual data are related to the current state of the robot and the input data acquired during the log-in and drinks orders from the totem. 

\subsubsection{User, Facial and Engagement Recognition Nodes}
Authentication mechanisms are widely used both in online and mobile systems. In particular, researchers have been conducting extensive efforts to create and improve authentication systems, such as biometric-based, that can be faster and easier to manage than password account management \cite{8590812}. The biometric-based authentication systems may use data such as voice, iris, fingerprint, palm-print, and face. %\cite{4813386,628669,10.1007/978-981-13-1135-2_42}. 
In a bartending scenario, multiple users access and complex interactions are expected, therefore the recognition of the user needs to be developed considering real-time constraints, such as disturbances caused by noisy and vast areas, the presence of many users, and the need for fast service to avoid crowds. The most used and appropriate technique for recognising users in similar scenarios is face recognition \cite{8590812}.
The identification, detection, and tracking of the users %by the agents involved in a BRILLO scenario 
are carried out using two cameras (an RGB camera at the totem, and a panoramic camera at the bar station) that allow the agent to collect data in real-time. The processed visual information is used for the following reasons: 1) face biometric data to identify a registered customer; 2) costumers' engagement via their body pose; 3) user tracking while at the counter; and 4) group recognition.

\begin{figure}[!t]
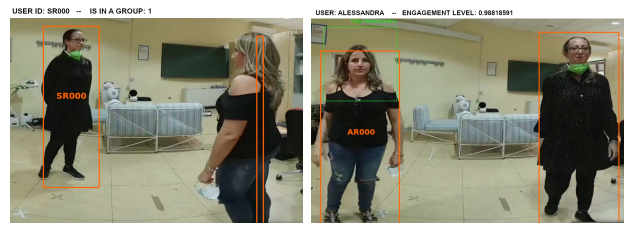

  \centering
  \includegraphics[width=.45\linewidth]{figures/group.png}
  \includegraphics[width=.48\linewidth]{figures/engagement.png}
  \caption{Screenshot of results from our engagement and group recognition modules: (a) Users in the figure on the left are recognised as in a group; (b) Users in the figure on the right are recognised as engaged with the robot.}
  \label{fig:group}
\end{figure}

The data from the camera are processed by YOLO\footnote{YOLO library \url{https://pjreddie.com/darknet/yolo/}} for object detection, while the face recognition is done through the OpenFace library\footnote{OpenFace library \url{https://github.com/TadasBaltrusaitis/OpenFace}}.
The combination of these two techniques allows accurate results even with different lights, lower quality of the frames and not fully frontal faces. The user pose is estimated using a Skeleton-Based approach through the OpenPose library\footnote{OpenPose library \url{https://github.com/CMU-Perceptual-Computing-Lab/openpose}}. The system estimates whether a user belongs to a group of other people using a Multilayer Perceptron classifier trained on the dataset called Ego-Group \cite{10.1007/978-3-319-16634-6_33} that is one of the few public datasets having a robot egocentric view of the surrounding space. Moreover, a Multilayer Perceptron model based on the user's pose and group information classifies, trained on the same dataset, the user's engagement with the robot. Screenshots from the used dataset with the results of the Engagement recognition module are shown in Figure \ref{fig:group}. 
On 5-fold cross-validation, an average accuracy of 94.33\% was achieved for engagement prediction and 97.12\% for group identification. %Moreover, the OpenFace library allows detecting the presence of multiple users, and the distance between the dominant face and the camera, which are useful information for building the priority queue of customers that the bartender robot needs to serve.

\subsubsection{Intent Recognition Node}
A bartending stand is typically found in noisy areas where a lot of people chat and, possibly, where music is playing. For this reason, to enable speech-based interaction, it is necessary to design the stand to be equipped with adequate hardware to isolate the customers' voices. This module makes use of the following different sub-processes: 1) automatic speech recognition (ASR) system to obtain the speech transcription, and 2) machine learning approaches for Natural Language Understanding (NLU).
 
The BRILLO bartender stand is equipped with two microphone arrays to perform source separation and noise reduction to isolate the customers' voices. The audio stream is processed remotely using an instance of the Azure ASR service, which produces the utterances' transcriptions.
 
Concerning the NLU module, the following seven intents are modelled using the Microsoft Azure Service LUIS\footnote{https://www.luis.ai/}:
 \begin{itemize}
     \item AnswerGreeting: this answers to the greeting phase initialised by the robot; this phase can also include information concerning the user's state.
     \item OrderConfirm: with this intent, the order is confirmed.
     \item OrderReject: with this intent, the order is conversely disconfirmed.
     \item DeleteOrder: the order is requested to be deleted.
     \item Help: with this intent, the user can ask for information about the possible actions the robot can perform.
     \item Menu: the user can request the available products.
     \item Order: the order is finalised or modified.
     \item NewsConfirm: the news proposed during the preparation phase is well received.
     \item NewsReject: the news proposed during the preparation phase is not well received.
     \item Evaluation: the evaluation concerning a previous order is processed.
 \end{itemize}
 
As far as orders are concerned, the user intents are modelled for the application domain, considering the type of product, possible modifications, and cancellations.

Intent performances were computed by dividing the example collected in the data set by subdividing them into a training set (80\%) and a test set (20\%).
The results are shown in Table \ref{tab:nlu-performances}. While for most intents, the F-score is high, for others (i.e., Menu, Help) the performance needs to be increased by adding further data. On average, the intent recognition module performed well with an F-score of 0.87.

 \begin{table}[]
\caption{Intent Recognition Performances}
\label{tab:nlu-performances}
% \scriptsize
 \centering
\begin{tabular}{|l|l|l|l|}
\hline
\textbf{Intent} & \textbf{Precision} & \textbf{Recall} & \textbf{F-score} \\ \hline
AnswerGreeting  & 1                  & 1               & 1                \\ \hline
OrderConfirm         & 1              & 1               & 1            \\ 
\hline
OrderReject     & 1              & 1          & 1             \\ 
\hline
Help            & 0.5                & 1               & 0.67             \\ \hline
Menu            & 1                  & 0.5             & 0.67             \\ \hline
Order           & 1                  & 0.81            & 0.9              \\ \hline
NewsConfirm    & 0.71               & 0.83               & 0.77             \\
\hline
NewsReject    & 0.86               & 0.75               & 0.80             \\
\hline
Evaluation &     1               & 1                  & 1                  \\
\hline
\end{tabular}
\end{table}

%\subsubsection{Sensors Filtering}
%This module selects the proper sensors to be activated in order to monitor the situation relying on the information provided by the Attentional Monitoring module. When there are multiple users, it acts as a filter by selecting from the same perceptual input the data of the user that is in the current focus of attention, as decided by the Attention Monitoring.  

%This filtering technique aims at preserving computational costs due to the processing of a huge amount of data, and reduce the noise occurring during parallel interaction or overlapping dialogues.

%% file: sections/beliefs.tex
According to Kotseruba and Tsotsos \cite{Tsotsos}, different types of memory can be identified in a cognitive architecture. 
In our architecture, a short-term memory stores the information related to the current users and situation states (e.g., current customers' engagement state, users' interaction states).% as processed by the deployed \textit{Human assessment} mechanisms. 

A working memory keeps track of the global state of the systems in terms of the list of orders to be served, that are represented in terms of goals to be reached, the current active intentions to be executed, and their plans of execution 
%that is continuously updated by the \textit{Execution layer}, and 
that are composed by both service and interactive actions. 

Long-term memory is used to store information about users' static information (personal data), preferences, and previous interactions' history. The users' personal data are stored in a MySQL database, while the length of the interaction, topics found of interest for the conversation, and interaction preferences, such as the type of the ordering (at the totem or at the bar), and an average estimation of the engagement are stored with the drinks orders. 

Finally, semantic memory is used to store semantic information (ontology-like structure) relying on the use of the Neo4J\footnote{\url{https://neo4j.com/}} graph database platform. The relationships between orders, cocktails and drinks, ingredients, and flavors are connected in a semantic graph that also includes the flavorDB\footnote{\url{https://cosylab.iiitd.edu.in/flavordb/}} database. The users' past interactions stored in the long-term memory are also associated with the semantic graphs.

%% file: sections/awareness.tex
This layer stores and processes the proper context-awareness information needed by the robot to produce socially acceptable and natural behaviour and to prepare the drinks. Moreover, such information is used to decide the user to be served/interacted with and the plan of actions.

\subsubsection{Turn-taking Node}
For each recognised user (\textit{/users} topic), the BRILLO system instantiates a desire for interaction and drink order. To accomplish such a goal, we defined a set of possible interaction states for the user, starting from the initial greetings towards the serving and farewell state. 
Each active user is currently in one state and, in order to transition from one state to another one, the proper set of actions (both service and interactive) have to be planned. 
The Turn-taking module selects one active user at the time (considering the arrival order, the waiting time, the presence of a group) and instantiates an intention for transitioning from one state to the following. 

\begin{figure}[!t]
  \centering
  \includegraphics[width=0.8\linewidth]{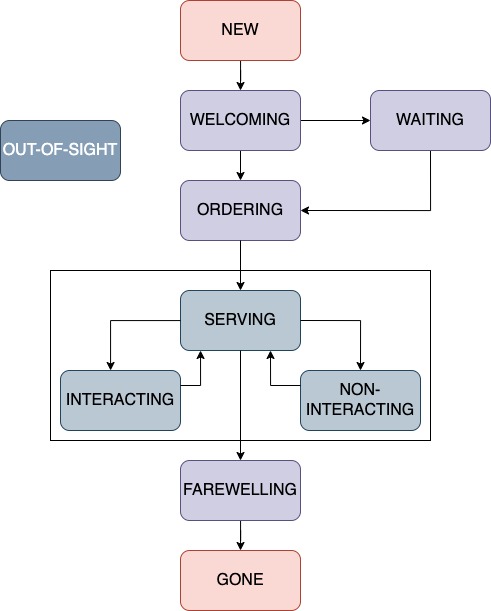}
  \caption{Possible interaction states for each user. Transitions between states are achieved by means of service and interaction actions. Users can transition from any state (except for ``gone") to an out-of-sight state, which will be re-assigned once the robot recognises them again.}
  \label{fig:userstates}
\end{figure}

However, based on the analysis of the users' associated personas and engagement level, some users will be observed by the robot as more involved in a conversation than others. Similarly, at any moment during the interaction, some users might claim the robot's attention. Notwithstanding these situations, the robot tries to involve all users, albeit to a different extent, in order to increase engagement \cite{1014396}.
%The robot's action for directing the gaze and starting a verbal interaction is guided by the beliefs (users and situation) provided by the \textit{Context-awareness and Decision-making} modules. Additionally, curiosity-driven approaches are used to provide the adaptive regulation of attention focus \cite{10.3389/fpsyg.2014.00273}.
Such user and their current interaction state are shared on the \textit{/interacting\_user} topic. States and transitions as represented in Figure \ref{fig:userstates}.

\subsubsection{Multi-modal Fusion Node}
The Multi-modal fusion module for the human assessment is deployed to 
%create data to populate the robot's memory through learning mechanisms. 
%Indeed, one of the principal skills required by the robot to 
keep the user's engagement in the interaction by recognising the user's intentional state and emotional reactions in order to decide which appropriate interactive action should be executed.   

Engagement and group detection information are here assessed also with 
emotion recognition from facial expressions that is carried out using the software called Affectiva\footnote{Affectiva software \url{https://www.affectiva.com/}}. The classification of people's emotions via Affectiva allows a classification of facial expressions according to seven main emotions (anger, contempt, disgust, fear, joy, sadness, and surprise). It also allows measuring the positive or negative valence for measuring the experience, with a frequency of a few hundred milliseconds and high accuracy.

The analysis of the voice, both simultaneous speech and high entropy, is implemented using the Python library called ~{My-Voice-Analysis\footnote{My-Voice-Analysis library \url{https://github.com/Shahabks/my-voice-analysis}}}. It allows evaluating people's mood (neutral, calm, or pacey), gender (female or male), speech rate, energy, frequency, average speech interval duration, and speaking duration. Moreover, in order to build a natural and fluid interaction, a social robot needs to be able to understand sentiments hidden in the content of the user's speech \cite{10.1007/978-3-319-98192-5_3}. Therefore, the user's voice is also further analysed to classify the emotions in the text. Currently, the sentiment analysis is carried out using the Azure Cognitive Services Text Analytics libraries\footnote{Azure services \url{https://azure.microsoft.com/}} which label the speech-to-text as positive, negative, or neutral.

The information collected here on the active user is used by the Interaction Manager to select the proper plan of interaction actions to keep the user engagement while transitioning to the following state.

%The emotions analysis based on facial, voice, and content of the text can be used to further model the choice of entertainment and casual dialogues to enhance the users' engagement and social interaction. 

\subsubsection{Robots' Actions Planner}
In our system, we distinguished two types of actions:
\begin{itemize}
    \item \textit{Service actions}, that are at the bartending actions necessary to prepare and serve a drink. It may involve one or both arms of the robot;
    \item \textit{Interactive actions}, that are the robot's actions for interacting and entertaining the user while they are at the bar station. These actions may be verbal utterances, gestures (when one of the robot's arms is not engaged in any service action), and facial expressions. 
\end{itemize}

\paragraph{Planner Node}
Orders (service actions) are processed by the robot as soon as the active user is in the serving state. ROS plan libraries for AI planning\footnote{ROSPlan's source code and documentation \url{https://github.com/KCL-Planning/ROSPlan}} are used to schedule service actions according to the scheduled orders for each user. The Planner module creates, therefore, a sequence of basic actions for each arm and for each ordered drink to be executed by the robot. The trajectories necessary to achieve each basic action to serve a drink are pre-recorded and basic actions are expressed in terms of preconditions to be checked before the execution (e.g., the mixer is empty) and the time to execute each action. 

\paragraph{Interaction Manager Node}

The selection of the interactive actions relies on Influence Diagrams\footnote{In BRILLO, Influence diagrams are implemented using the AGRUM library\cite{Hal2020} \url{https://agrum.gitlab.io/}}, which integrate the probabilistic estimates of engagement coming from the users' interaction modes (i.e. speech, facial expressions, semantics, pose) with a utility estimation linked to the possible speech acts or gesture. Bayesian networks allow handling probabilistic input coming from the NLU module, among others, to take into account confidence measures when selecting the next action. %The Bayesian Network managing probabilistic input provided by the NLU module %is shown in Figure \ref{fig:bayes}
%: while 
The utility of each possible machine move (Action) is a function of the chosen Action and of the actual intent from the user, at decision time only an estimate of the actual intent is available. The data summarised in Table \ref{tab:nlu-performances} inform the network of the probability of each intent to be correct and also assigns a utility value to the dialogue move consisting of a request for repetition (AskRepeat). Depending on the probability distribution over all intents, provided by the NLU module, the utility of each possible Action is computed. The Action with the highest utility is, then, executed. 

%\begin{figure}
%    \centering
%    \includegraphics[width=.5\linewidth]{figures/BayesianNet.png}
%    \caption{A Bayesian network managing uncertain input provided by the NLU module.}
%    \label{fig:bayes}
%\end{figure}

Interaction customisation is applied in this phase when the user is known to the system.
Pragmatic models of interaction are investigated and included in the architecture. For instance, clarification requests (CRs) expressing counter-expectations with respect to past ordering actions are adapted to engage the user in the dialogue and add further details to the user model.
For order confirmation, CRs with a confirmation function are employed, i.e. when the CR initiator has some kind of hypothesis \cite{rieser2005implications}. Specifically, this is useful when low confidence scores occur due to background noise or multiple orders. 

\begin{figure}
\centering
        \includegraphics[width=1\linewidth]{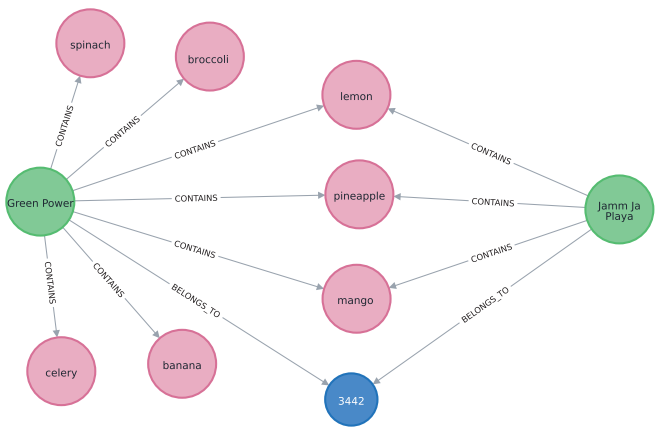}
    \caption{Graph showing the similarity between the last ordered DRINK node, \emph{Green Power}, and another DRINK node (in green) belonging to the same category SMOOTHIE (in blue). The similarity is here computed on the highest number of FOOD\_INGREDIENT nodes (in pink) in common.}
    \label{fig:similarity}
\end{figure}

To support recommendations, previous interactions with the users are stored in the Neo4j database and used to build a profile. For past interactions involving drinks that were not previously selected, the robot asks for explicit feedback, in the form of a rating on a scale of 1 to 5. Using the graph structure represented in Neo4j and, in particular, the data coming from FlavorDB (see Figure \ref{fig:similarity}), the system is able to compute drinks' similarity in terms of shared ingredients. On this basis, different recommendation strategies can be selected according to different parameters: a) type of persona, a not mandatory piece of information asked at the totem kiosk during the registration phase; b) degree of knowledge of the user based on the number of interactions (known user: at least one previous interaction; unknown user: first interaction); c) evaluation of past consumed drinks (positive evaluation: $\geq 3$; negative evaluation: $\leq 2$); d) acceptance of the recommendation (if the recommendation is not accepted another strategy is selected) (see Table \ref{tab:recommandation}).

\begin{table*}[t]
\centering
\caption{Recommendation Strategies}
%\normalsize
\begin{tabular}{|l|lll|}
\hline
\multicolumn{1}{|c|}{\textbf{Persona}}                                              & \multicolumn{3}{c|}{\textbf{Recommendation Strategies}}                                                                                                                                                                                                                               \\ \hline
\multirow{5}{*}{\textit{worker}}                                                             & \multicolumn{1}{l|}{\multirow{3}{*}{\begin{tabular}[c]{@{}l@{}}known\\ user\end{tabular}}} & \multicolumn{2}{l|}{a) client's preferred drink}                                                                                                                                         \\ \cline{3-4} 
                                                                                    & \multicolumn{1}{l|}{}                                                                      & \multicolumn{2}{l|}{b) users' most ordered drink}                                                                                                                                        \\ \cline{3-4} 
                                                                                    & \multicolumn{1}{l|}{}                                                                      & \multicolumn{2}{l|}{c) ask what the client wants to order}                                                                                                                               \\ \cline{2-4} 
                                                                                    & \multicolumn{1}{l|}{\multirow{2}{*}{new user}}                                             & \multicolumn{2}{l|}{a) users' most ordered drink}                                                                                                                                        \\ \cline{3-4} 
                                                                                    & \multicolumn{1}{l|}{}                                                                      & \multicolumn{2}{l|}{b) ask what the client wants to order}                                                                                                                               \\ \hline
\multirow{8}{*}{\begin{tabular}[c]{@{}l@{}}\textit{other}\\ (specified or not)\end{tabular}} & \multicolumn{1}{l|}{\multirow{6}{*}{\begin{tabular}[c]{@{}l@{}}known\\ user\end{tabular}}} & \multicolumn{1}{l|}{\multirow{4}{*}{last drink was positively evaluated}} & \begin{tabular}[c]{@{}l@{}}a) similar drink from the same category\\     (smoothie or cocktail)\end{tabular} \\ \cline{4-4} 
                                                                                    & \multicolumn{1}{l|}{}                                                                      & \multicolumn{1}{l|}{}                                                     & b) similar drink from another category                                                                       \\ \cline{4-4} 
                                                                                    & \multicolumn{1}{l|}{}                                                                      & \multicolumn{1}{l|}{}                                                     & c) users' most ordered drink                                                                                 \\ \cline{4-4} 
                                                                                    & \multicolumn{1}{l|}{}                                                                      & \multicolumn{1}{l|}{}                                                     & d) ask what the client wants to order                                                                        \\ \cline{3-4} 
                                                                                    & \multicolumn{1}{l|}{}                                                                      & \multicolumn{1}{l|}{\multirow{2}{*}{last drink was negatively evaluated}} & a) users' most ordered drink                                                                                 \\ \cline{4-4} 
                                                                                    & \multicolumn{1}{l|}{}                                                                      & \multicolumn{1}{l|}{}                                                     & b) ask what the client wants to order                                                                        \\ \cline{2-4} 
                                                                                    & \multicolumn{1}{l|}{\multirow{2}{*}{new user}}                                             & \multicolumn{2}{l|}{a) users' most ordered drink}                                                                                                                                        \\ \cline{3-4} 
                                                                                    & \multicolumn{1}{l|}{}                                                                      & \multicolumn{2}{l|}{b) ask what the client wants to order}                                                                                                                               \\ \hline
\end{tabular}
\label{tab:recommandation}
\end{table*}

%Influence Diagrams are also used to manage the robot's interactive action depending on an estimate of the usefulness of social interactions using Bayesian learning. This includes recommending a new product or providing a topic of conversation. User profiles can be, then, modelled in terms of probability distributions predicting social feedback for each interactive actions sequence based on previous interactions. Utility, for each choice, is computed using positive social feedback from the customers as a reward function. The system obtained the positive social feedback from the customers as both online process and afterwards. For example, the robot asks if the customer enjoyed a news or information it provided while engaging them in small talks.

For entertainment purposes, the robot is also able to present news extracted daily from different sources and categories. Explicit feedback is used in this case, too, to support user profiling and to select topics from the most appropriate categories. More specifically: first the robot suggests a topic of interaction, based on similarities among users and/or personas; once a topic is accepted, the corresponding news is selected from a serious or entertaining source; afterwards, the user is asked to give feedback concerning their interest about the news. If the feedback is positive, another news belonging to the same category is proposed, otherwise the category is changed. News items can be provided as long as available. Other social signals can also cause the robot to stop entertaining the current client, i.e. when the client is not engaged or interested, or when another registered client appears in the robot's visual area.

%% file: sections/exec.tex
The \textit{Execution layer} manages the orders and the interactions of the robots with the users according to the knowledge and beliefs of the agents, and the situational context.

\subsubsection{Action Manager Node}

The Action Manager module acts as a scheduling mechanism to improve the efficiency in making the drinks by using one or two arms in parallel.
%, and the human and situation awareness (e.g., the engagement state of the users). 
It schedules the interactive actions, related to the robot's gestures, and the service actions by orchestrating the robot's arms movements. % \cite{conf/ro-man/CaccavaleLLRSF14}. %This allows from time to time to change the focus of attention form one customer toward another one. 
%Interactive actions are activated in response to both scheduled orders and context-awareness. Meaning that the robot can socially interact with a customer who is not currently served. 
The next actions are chosen to balance the expected contribution of the action towards the goal of preparing a drink, and the robot's actions needed to maintain the engagement and entertainment of the users to their social expectations. For example, the robot might engage the users in a casual conversation while preparing their drink for fostering a more endearing interaction %. The robot may also adapt its body posture making gestures 
if its arms are not busy in the drink preparation. %The robot may decide to lead the conversation toward the end, if the users finished their drink and other customers are waiting.
%oriented to different customers at the bar also depending on other factors () 
%(in the case of a customer waiting at the table) or when the user is identified by the robot (in the case of a customer served at the bar). 

 %\cite{Hal2020}.
%Dynamic action selection involves the selection of one interactive action toward a single user based on knowledge at the time in a probabilistic selection mechanism. 
%This is implemented using probabilistic graphical models applied to the AGRUM library \cite{Hal2020} in the form of Influence Diagrams. These diagrams 

%The communications between these modules and the \textit{Context-awareness layer} are bidirectional to allow the robot to dynamically plan actions (single or multiple), and behaviours according to the user and the situational context.  This module collects the users' \textit{desires} of some actions from the dialogue manager, and decides whether the action planner and controller should transform it into a \textit{goal} and, then, generates \textit{intentions}.

\subsubsection{Arms Node}
This module is a wrapper for favouring the communication between the BRILLO system and the two KUKA arms' Programmable Logic Controller. The controller has been implemented using the KUKA SunriseOS system to securely manage the arms, the grippers and each movement in the stand.

\subsubsection{Furhat Interface Node}

%Intrinsic motivations for action selection? drives? (mariacarla?)
%Turn-taking is managed considering both top-down and bottom-up stimuli. 
To make the interaction more engaging, the robot is endowed with a module in charge of guiding the Turn-taking through gaze and speech pauses \cite{Mutlu09}.
The system, in particular, manages the behaviours of this robot by considering the presence and the absence of the user's attention. When a low user's engagement is perceived, the robot produces facial expressions and vocal sounds to catch the user's attention. During a dialogue, the robot adapts its facial expressions and vocal sounds whether it is listening to the user's speaking, it understood or did not understand the speech, and to express an emotion in relation to the topic of conversation. The facial and vocal sounds are based on the Facial Action Coding System (FACS) for Ekman's emotions\footnote{FACS \url{shorturl.at/fvGP4}} and implemented using Python and Furhat Robotics' APIs. 
%Priority will be given to the users depending on the chronological sequences of orders. 

%% file: sections/conc.tex
The use of robots for the automation of the supply of food and beverages is a commercially attractive and modern application of robotic technologies and it is used as a strategy to renew the image of the service and thus stimulate people's curiosity. However, the maintenance, in the long term, of the degree of interest in a commercial proposal linked to leisure can be only effectively achieved by implementing strategies for personalising the experience according to previous interactions and adapting the robot's behaviours. % to the user's needs. 

In this work, we presented a ROS Architecture for a bartending robot aiming at providing an efficient service while keeping engaged the human customers in a dynamic and personalised interaction. The system has been tested in a controlled environment evaluating the effectiveness of the single developed modules and of the whole architecture in terms of personalization of the experience and adaptation of behaviours, including content and frequency of dialogues, movements and poses to the customers' moods and preferences. Software modules were designed for the management of dialogues and the perception of social signals, such as, for example, biometric-based algorithm for the recognition of social signals linked to the dynamics of interaction with the robot and Bayesian networks for the modelling of the interaction and the estimation of the probability of success of new proposals based on previous interactions. 

As soon as the pandemic restriction will allow it, the system will be tested in an ecological environment to understand if BRILLO's robots are able to meet also people's social expectations. 
We aim at demonstrating that receiving a service that is fast but also personalised can present an added value to the commercial proposal, taking it beyond the simple experience associated with the presence of a new technology and increasing customers' loyalty. 